\documentclass{article}

\usepackage[final,nonatbib]{nips_2017}

\usepackage[utf8]{inputenc} 
\usepackage[T1]{fontenc}    
\usepackage{hyperref}       
\usepackage{url}            
\usepackage{booktabs}       
\usepackage{amsfonts}       
\usepackage{nicefrac}       
\usepackage{microtype}      

\usepackage{graphicx,xcolor}

\title{Revisiting hand-crafted feature for action recognition: a set of improved dense trajectories}

\author{
  Kenji Matsui, Toru Tamaki \\
  Hiroshima University, Japan\\
  \And
  Gwladys Auffret \\
  ENSICAEN, France \\
  \And
  Bisser Raytchev, Kazufumi Kaneda\\
  Hiroshima University, Japan
}

\begin{document}

\maketitle

\begin{abstract}
We propose a feature for action recognition called Trajectory-Set (TS),
on top of the improved Dense Trajectory (iDT).
The TS feature encodes only trajectories around densely sampled interest points,
without any appearance features.
Experimental results on the UCF50, UCF101, and HMDB51 action datasets
demonstrate that TS is comparable to state-of-the-arts, and outperforms many other methods;
for HMDB the accuracy of 85.4\%, compared to the best accuracy of 80.2\% obtained by a deep method.
Our code is available on-line at 
\url{https://github.com/Gauffret/TrajectorySet}.
\end{abstract}

\section{Introduction}
\label{sec:introduction}

Action recognition has been well studied in the computer vision literature \cite{Herath2017}
because it is an important and challenging task.
Deep learning approaches have been proposed recently \cite{Simonyan2014,Feichtenhofer2016,Wang2016},
however still a hand-crafted feature, improved Dense Trajectory (iDT)
\cite{DT,iDT}, is comparable in performance. Moreover,
top performances of deep learning approaches are obtained
by combining the iDT feature \cite{Feichtenhofer2016,Tran2015,Wang2015}.

In this paper, we propose a novel hand-crafted feature for action recognition,
called Trajectory-Set (TS), that encodes trajectories in a local region of a video
\footnote{This work has been published in part as \cite{Matsui2017}}.
The contribution of this paper is summarized as follows.
We propose another hand-crafted feature that can be combined with deep learning approaches.
Hand-crafted features are complement to deep learning approaches, however
a little effort has been done in this direction after iDT.
Second, the proposed TS feature focuses on the better handling of motions in the scene.
The iDT feature uses trajectories of densely samples interest points in a simple way,
while we explore here the way to extract a rich information from trajectories.
The proposed TS feature is complement to appearance information such as HOG and objects in the scene,
which can be computed separately and combined afterward in a late fusion fashion.

There are two relate works relevant to our work.
One is trajectons \cite{Matikainen2009} that uses a global dictionary of trajectories in a video
to cluster representative trajectories as snippets. Our TS feature is computed locally,
not globally, inspired by the success of local image descriptors \cite{HOG}.
The other is the two-stream CNN \cite{Simonyan2014} that uses a single frame and a optical flow stack.
In their paper stacking trajectories was also reported but did not perform well,
probably the sparseness of trajectories does not fit to CNN architectures.
In contrast, we take a hand-crafted approach that can be fused later with CNN outputs.

\section{Dense Trajectory}

Here we briefly summarize the improved dense trajectory (iDT) \cite{iDT}
on which we base for the proposed method.
First, the image pyramid for a particular frame at time $t$ in a video is constructed,
and interest points are densely sampled at each level of the pyramid.
Next, interest points are tracked in the following $L$ frames ($L=15$ by default). 
Then, the iDT is computed by using local features such as 
HOG (Histogram of Oriented Gradient) \cite{HOG}, 
HOF (Histogram of Optical Flow), and 
MBH (Motion Boundary Histograms) \cite{MBH}
along the trajectory tube; a stack of patches centered at the trajectory in the frames.

For example, between two points in time $t_0$ and $t_L$,
a trajectory $T_{t_0, t_L}$ has points $p_{t_0}, p_{t_1}, \ldots, p_{t_L}$
in frames $\{t_0, t_1, \ldots, t_L\}$.
In fact, $T_{t_0, t_L}$ is a vector of displacement between frames rather than point coordinates, that is,
$T_{t_0, t_L} = (v_0, v_1, \ldots, v_{L-1})$
where $v_i = p_{i-1} - p_i$.
Local features such as $\mathit{HOG}_{t_i}$ are computed with a patch centered at $p_{t_i}$ in frame at time $t_i$.

To improve the performance, the global motion is removed by computing homography,
and background trajectories are removed by using a people detector.
The Fisher vector encoding \cite{FisherVector} is used to compute an iDT feature of a video.

\begin{figure}[t]
\centering
\includegraphics[width=\linewidth]{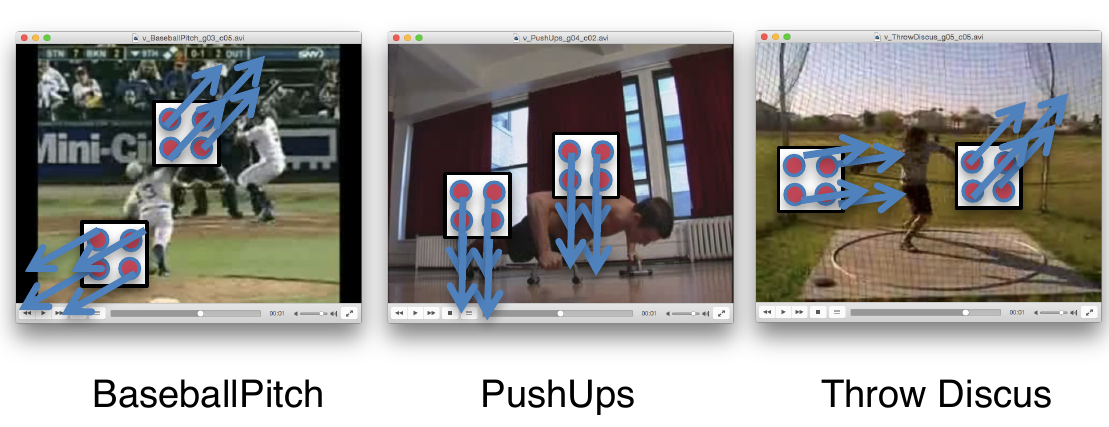}
\caption{Different actions in UCF50 \cite{UCF50} have different trajectory information.}
\label{fig:different_categories}
\end{figure}

\begin{figure}[t]
\centering

\begin{minipage}{.35\linewidth}\centering
\includegraphics[width=\linewidth]{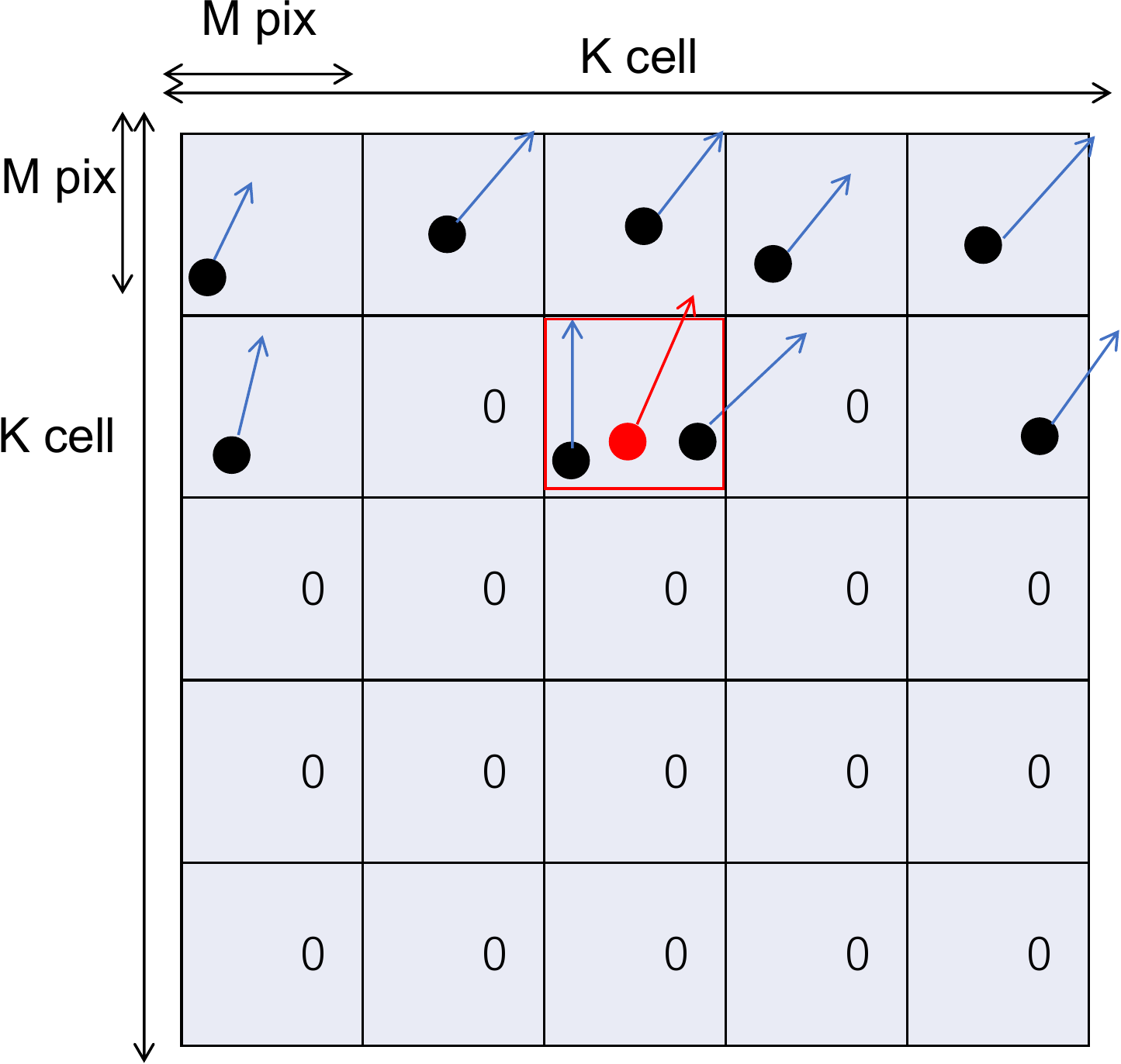}\\
(a)
\end{minipage}
\begin{minipage}{.6\linewidth}\centering
\includegraphics[width=.8\linewidth]{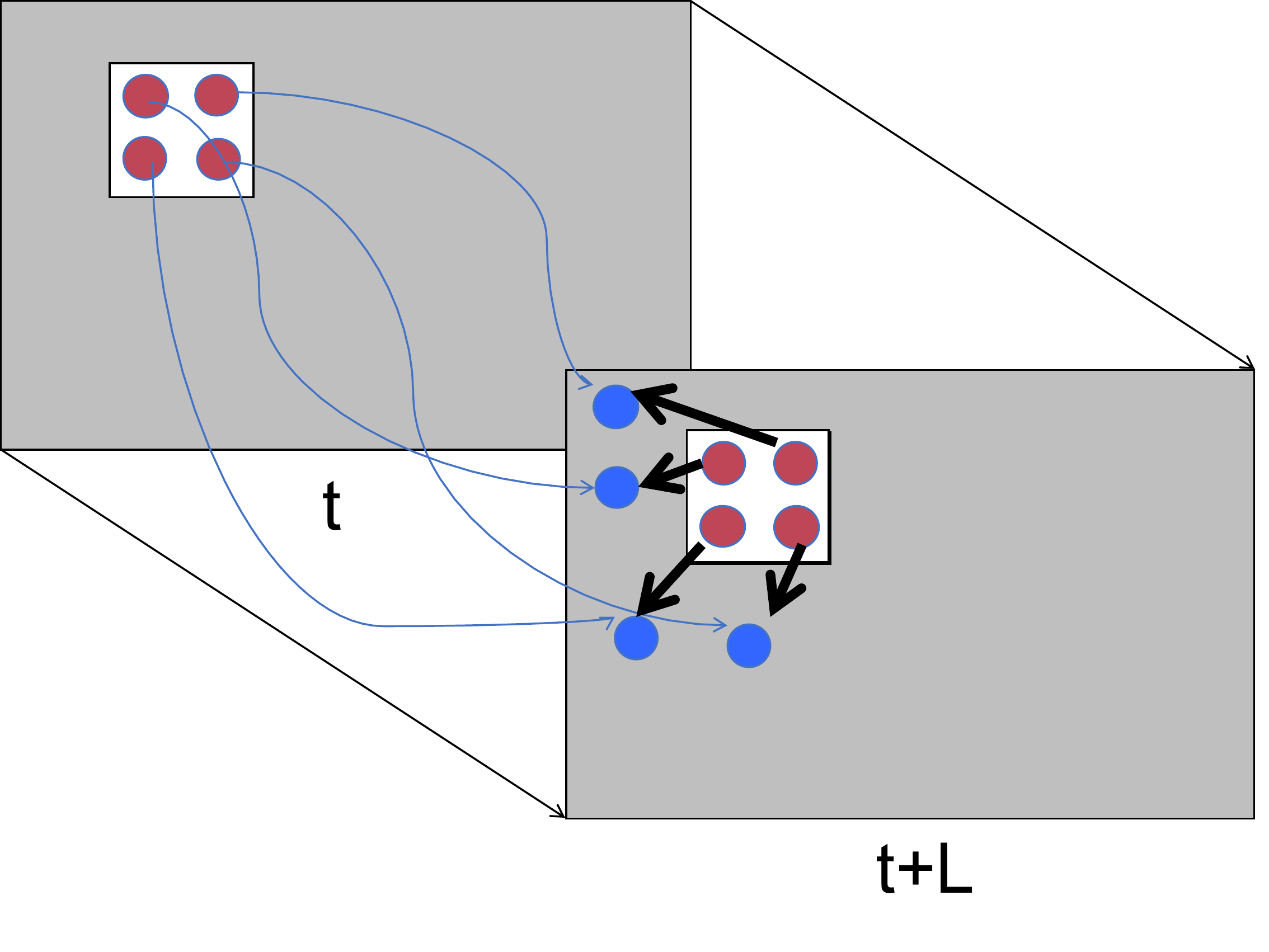}\\
(b)
\end{minipage}

\caption{
(a) 
A block and cells in the starting frame.
Starting points of trajectories in each cell are shown in black circles with motion vector arrows.
Cells with no starting points are filled with 0.
If there are multiple trajectories starting from the same cell, 
the average trajectory is used for the cell (the averaged starting point is shown in red in this figure).
(b)
A Trajectory-Set feature consists of $K^2$ trajectories (shown as blue curves)
starting from the same block in the starting frame $t$ and wander across the successive $L$ frames.
Magenta circles are the starting points of trajectories,
and blue circles are corresponding end points.
The displacement vectors between starting and end points are shown as black arrows.
}

\label{fig:block_cells}
\label{fig:trajectory-set}

\end{figure}

\section{Proposed Trajectory-Set feature}

We think that extracted trajectories might have rich information discriminative enough for classifying
different actions, even although trajectories have no appearance information.
As shown in Figure \ref{fig:different_categories},
different actions are expected to have different trajectories,
regardless of appearance, texture, or shape of the video frame contents.
However a single trajectory $T_{t_0, t_L}$ may be severely affected
by inaccurate tracking results and an irregular motion in the frame.

We instead propose to aggregate nearby trajectories to form a Trajectory-Set (TS) feature.
First, a frame is divided into non-overlapping cells of $M \times M$ pixels as shown in Figure~\ref{fig:block_cells}(a).
Next, $K \times K$ cells form a block%
\footnote{Note that we borrow the terms from HOG \cite{HOG}.}.
This results in overlapping blocks of $MK \times MK$ pixels with spacing of $M$ pixels.


The key concept of the TS feature is to collect trajectories
that start in a local region (or block) in the starting frame
(see Figure \ref{fig:trajectory-set}(a)).
In each cell of a block in the starting frame,
we find a trajectory starting from the cell.
(If there are multiple trajectories starting from the cell, the average trajectory is used.
If no trajectory starts from the cell, we use a zero vector as the trajectory of the cell.)
By repeating this procedure for all $K\times K$ cells in the block,
we have a set of trajectories starting from the block.
We concatenate the trajectories to form a TS feature of dimension $2L K^2$ for the block.
As shown in Figure \ref{fig:trajectory-set}(b), the TS feature consists of trajectories
that start in the same block in the starting frame and wander across frames.
Note that the end points of the trajectories are not necessary close to each other.
This implies that we enforce the locality of trajectories only in the starting frame.

\begin{figure}[t]
\centering

\begin{minipage}{.32\linewidth}\centering
\includegraphics[width=.8\linewidth]{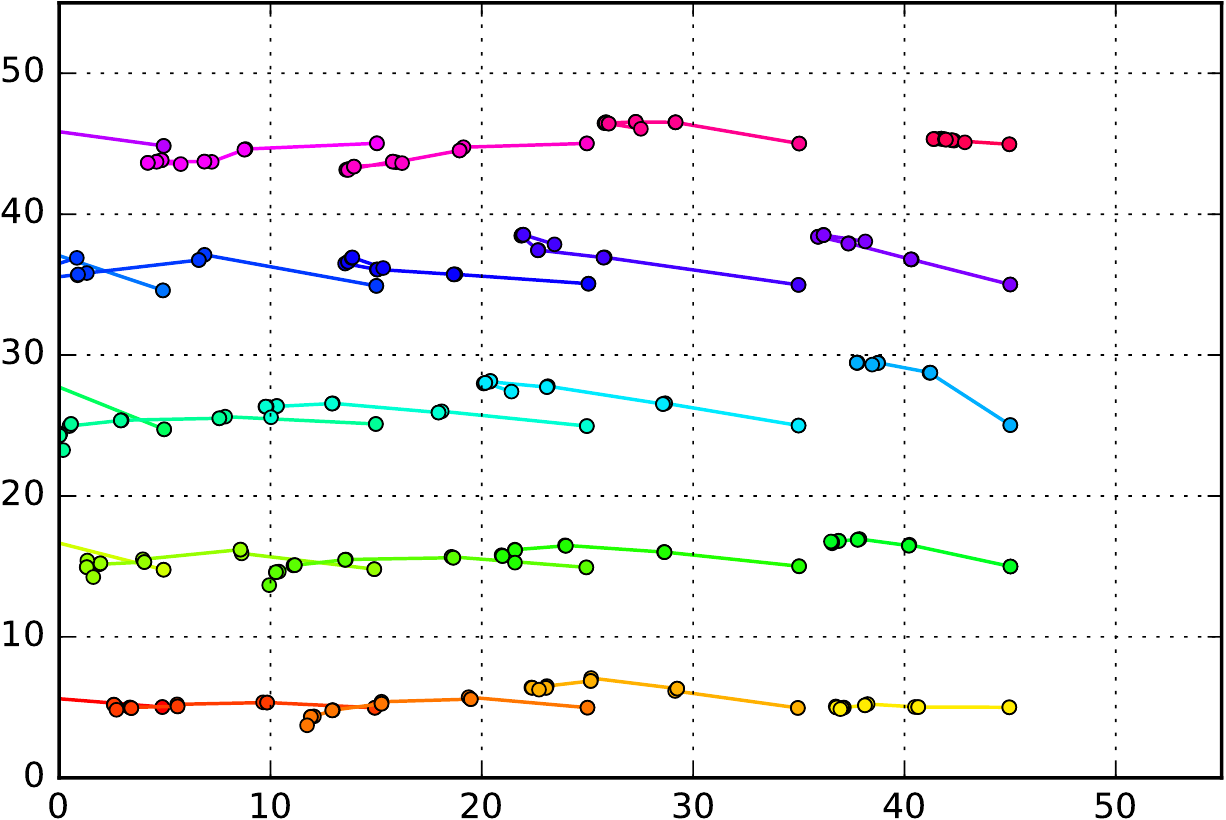}\\
\includegraphics[width=.8\linewidth]{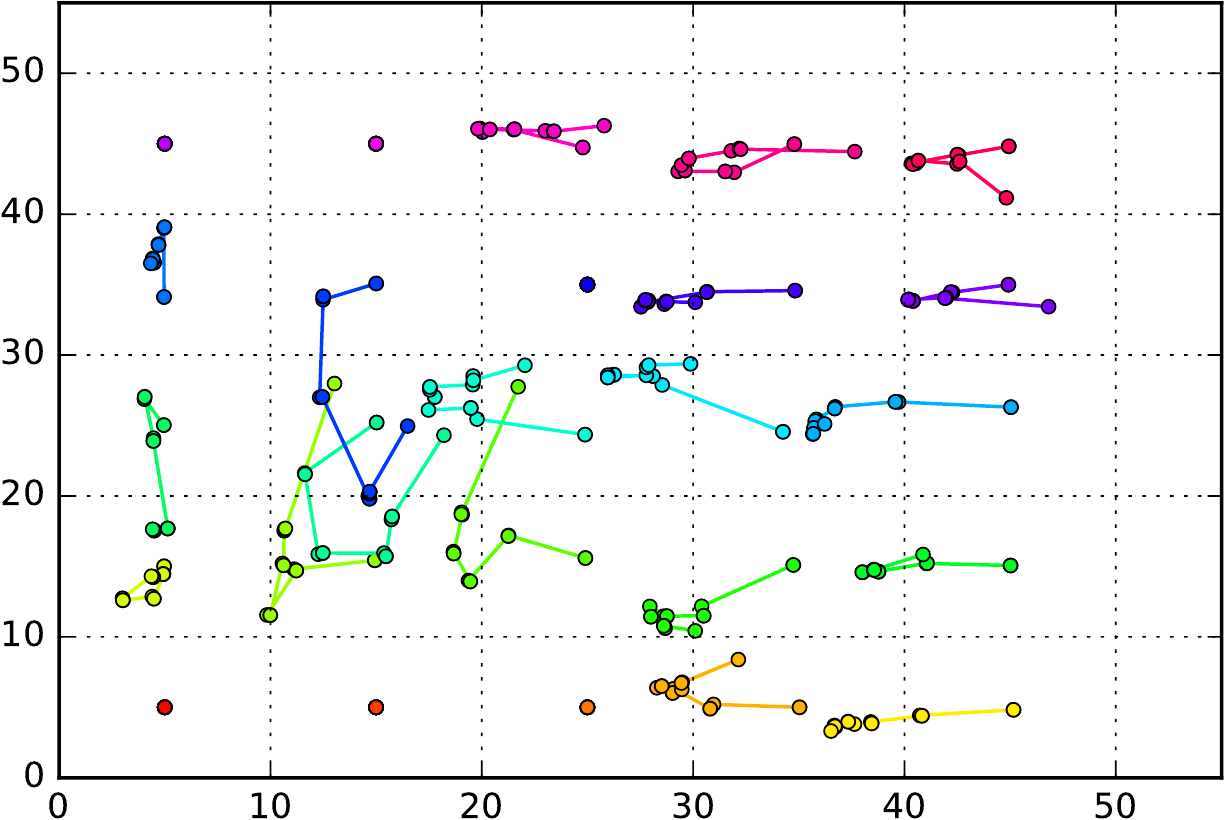}\\
\includegraphics[width=.8\linewidth]{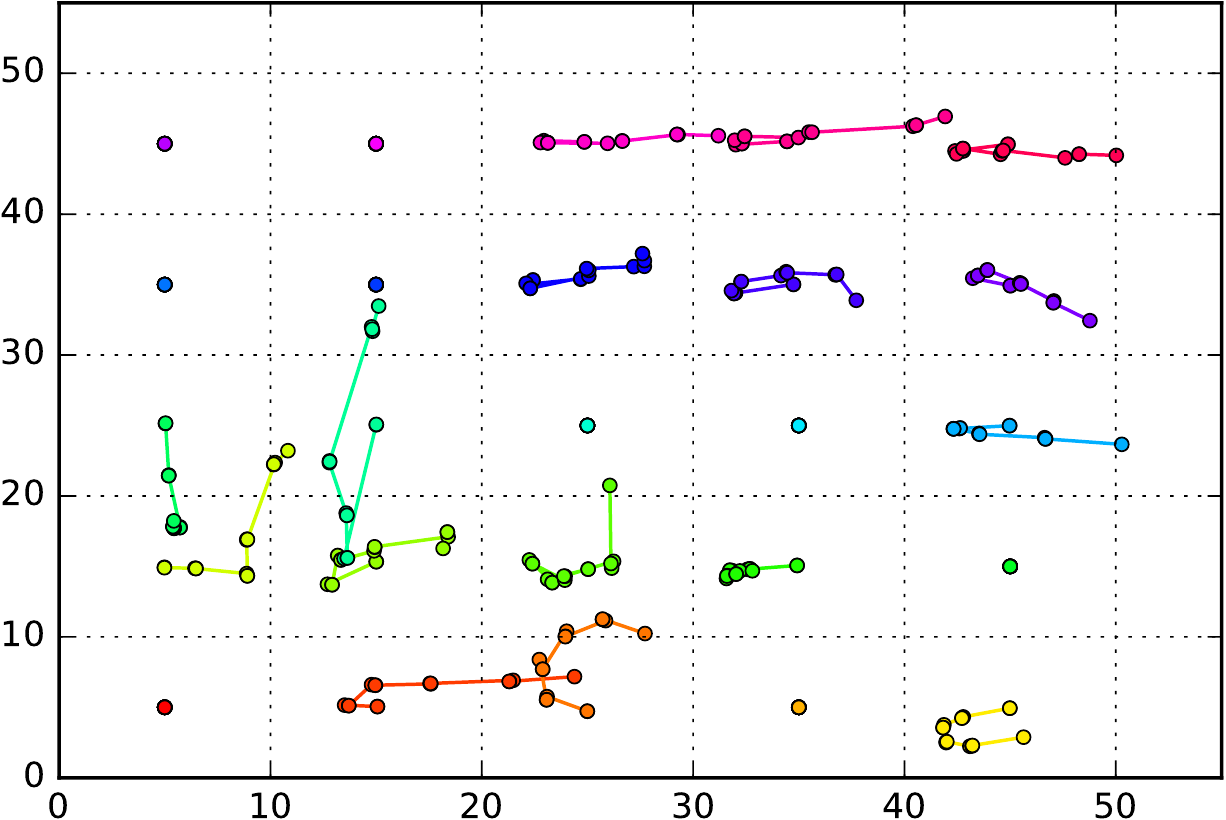}\\
(a)
\end{minipage}
\hfill
\begin{minipage}{.32\linewidth}\centering
\includegraphics[width=.8\linewidth]{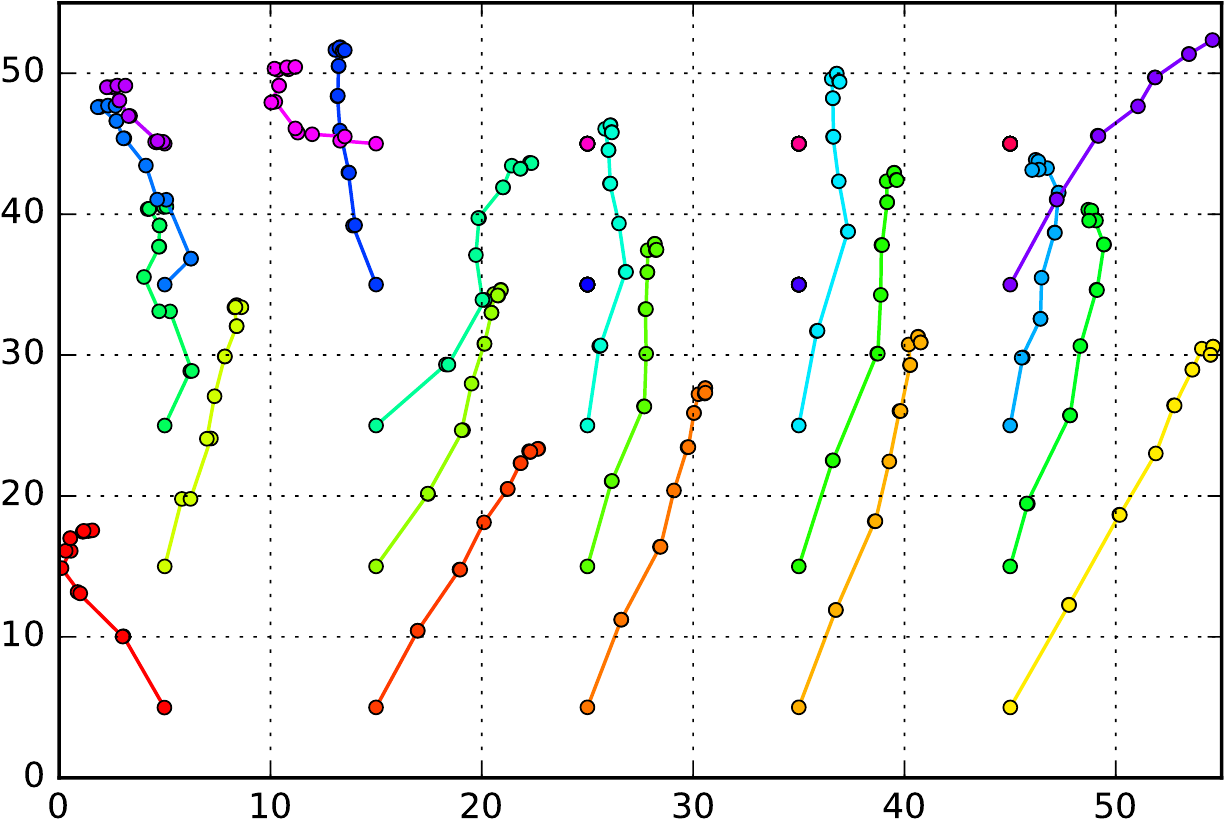}\\
\includegraphics[width=.8\linewidth]{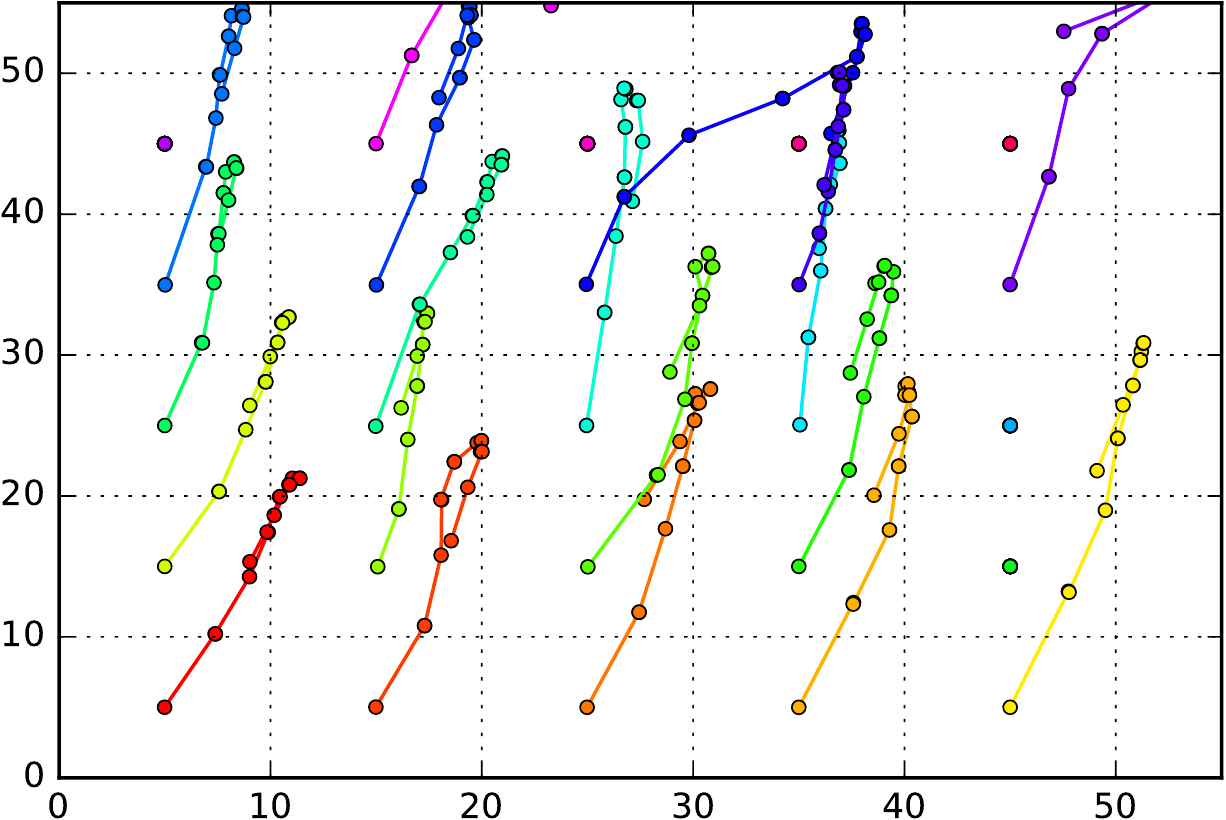}\\
\includegraphics[width=.8\linewidth]{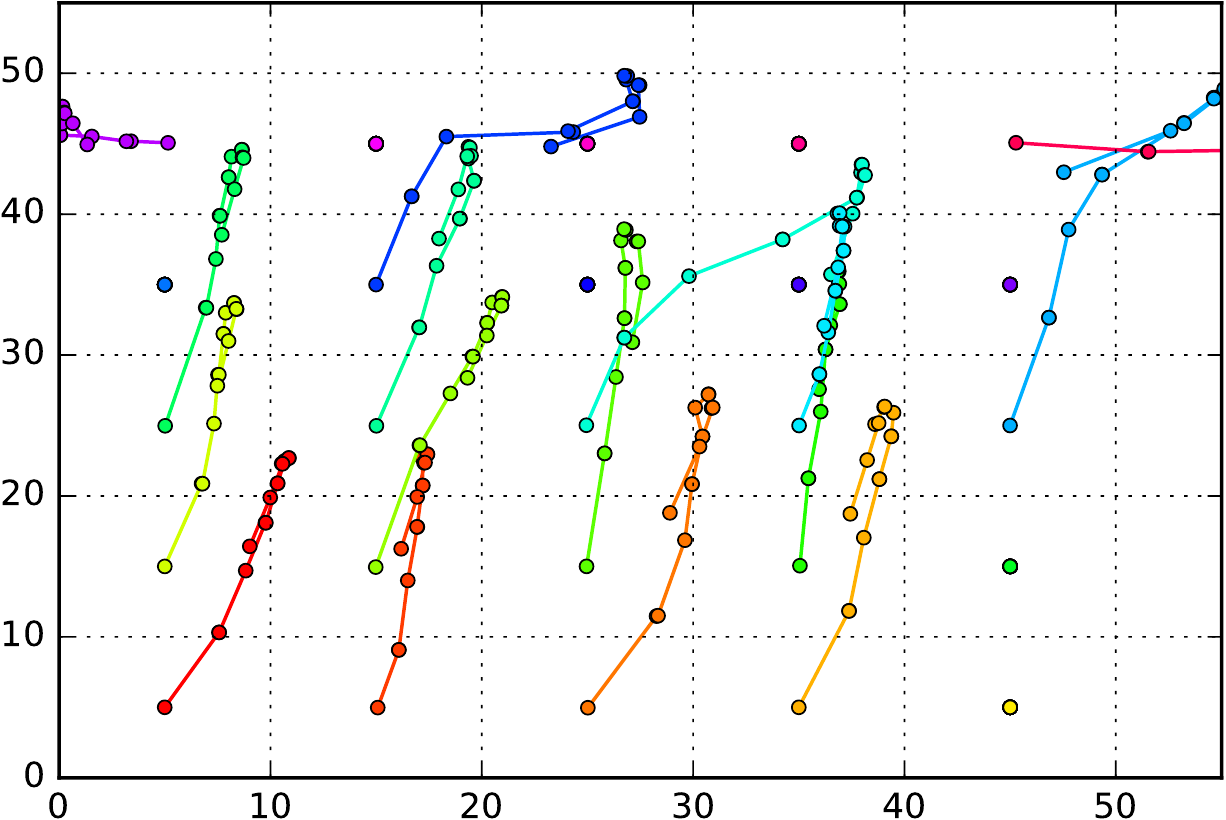}\\
(b)
\end{minipage}
\hfill
\begin{minipage}{.32\linewidth}\centering
\includegraphics[width=.8\linewidth]{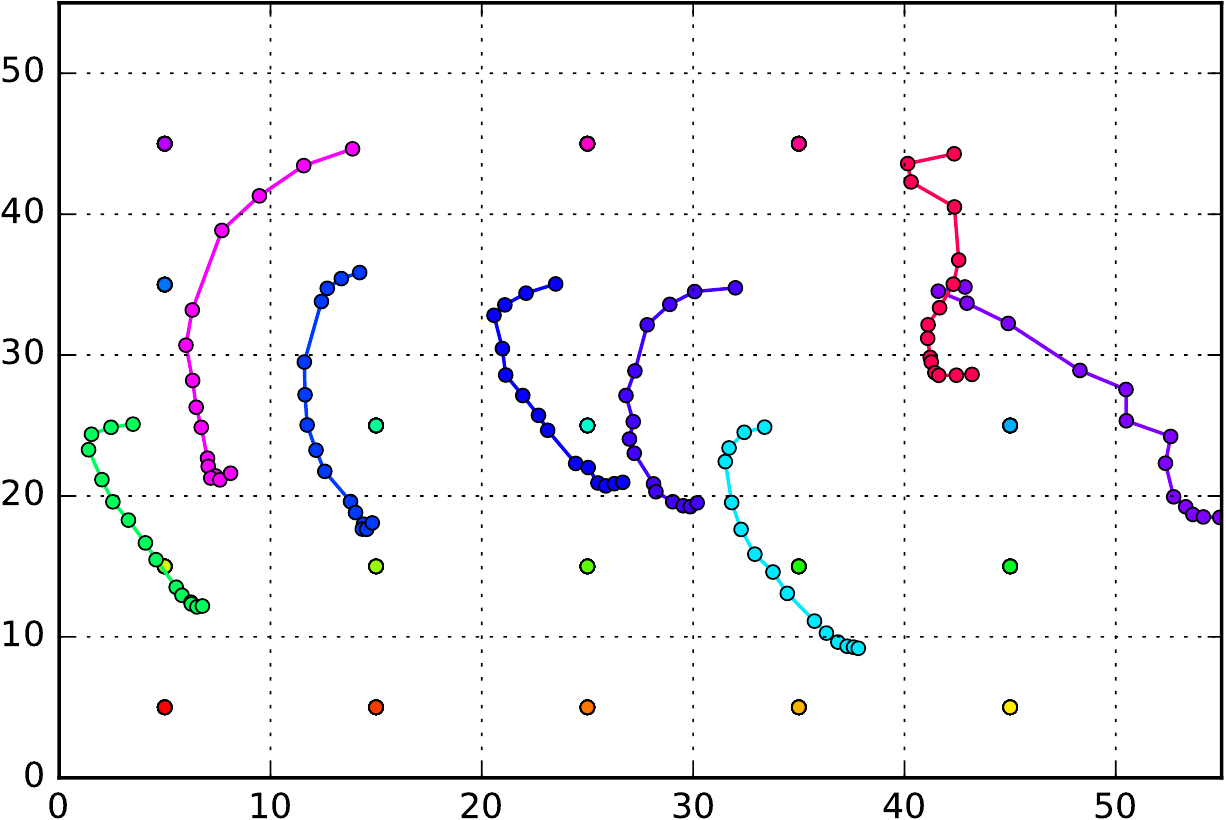}\\
\includegraphics[width=.8\linewidth]{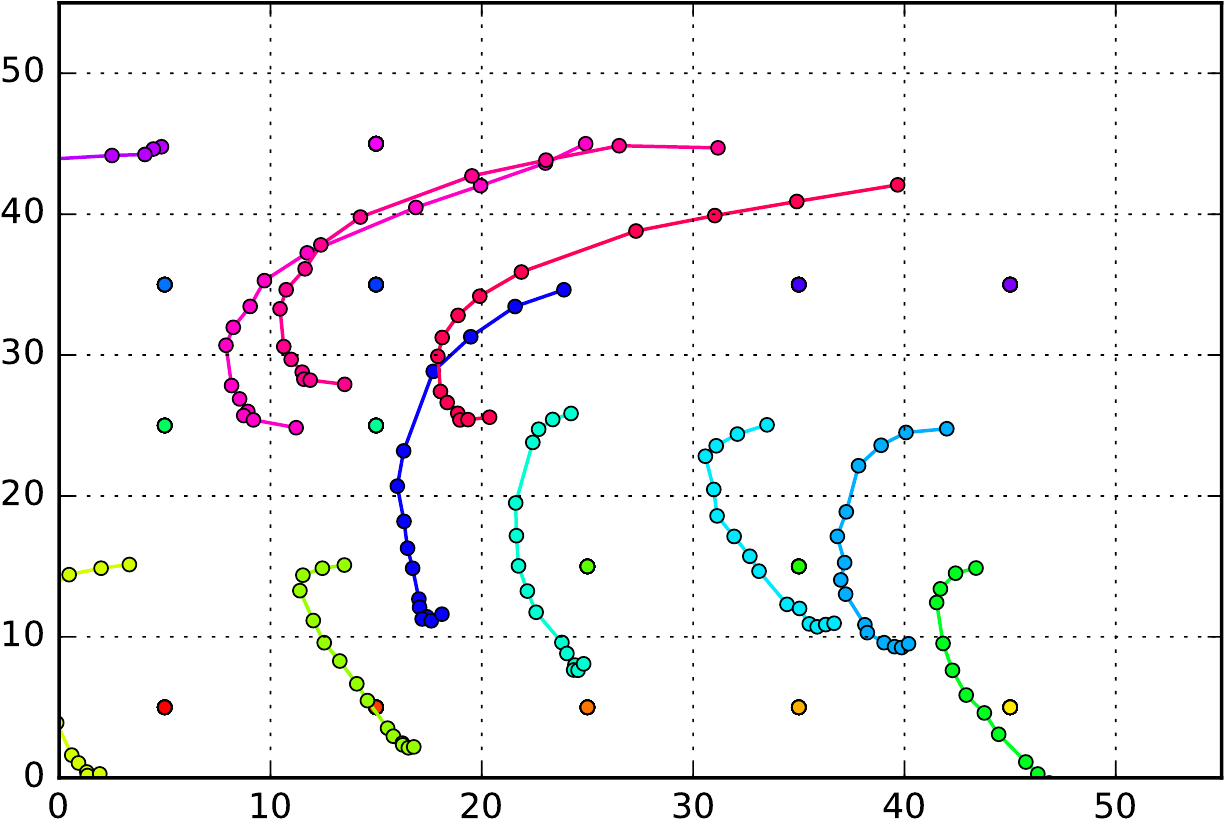}\\
\includegraphics[width=.8\linewidth]{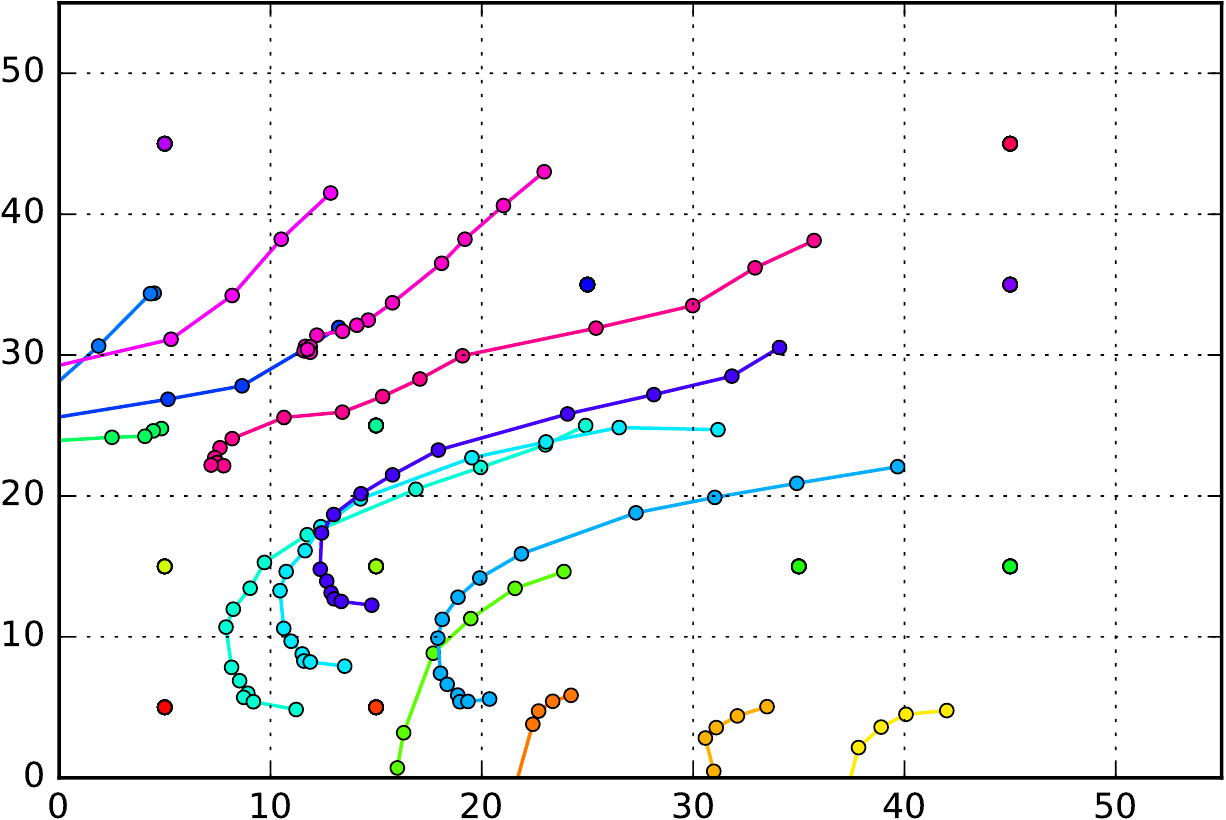}\\
(c)
\end{minipage}

\caption{Examples of TS features of (a) BaseballPich, (b) PushUps, and (c) ThrowDiscus in the UCF50.
Each row shows different TS features obtained from different blocks and different sets of 15 frames.
Each plot shows 25 trajectories (in different colors) starting from each of cells in a block.
Trajectories are shown with 16 points (some points are overlapped) connected with lines.
The block and cell sizes are $50 \times 50$ and $10 \times 10$ pixels, respectively.
}
\label{fig:raw_trajectories}
\end{figure}

In our default setting, $L=15$, $M=10$, and $K=5$, then the TS feature is a 750 dimensional vector.
Figure \ref{fig:raw_trajectories} shows examples of TS features for different categories.
We can see different motion patterns appear in each of TS features.

Here we can propose some variations.
Instead of using a trajectory as a series of displacements
$T_{t_0, t_L} = (v_0, v_1, \ldots, v_{L-1})$,
we can simply a series of coordinates like as $T_{t_0, t_L} = (p_0, p_1, \ldots, p_L)$,
but in local coordinate systems instead of the global coordinate system.
For further reducing computation cost,
we can skip every two frames by summing successive two displacement vectors
(that is, by skipping one frame in $(v_0, v_1, \ldots, v_{L-1})$ to generate $(v_0+v_1, v_2+v_3, \ldots)$),
resulting in feature vectors of dimension 400.
We call these processes ''skip2'' in the results.

\section{Experimental results and discussion}

Here we describe experimental results of the proposed method.
We used UCF50 \cite{UCF50}.
It has 50 action categories.
Videos in each category are divided into 25 groups, 
and we evaluate the accuracy with the leave-one-group-out cross validation.
The resolution of videos are $320 \times 240$ @ 30fps,
and the durations are between 1 and 6 seconds.
For TS feature construction, we use $M=10$ pixels, $K=5$, and $L=15$,
and randomly sample 1\% of TS features for encoding with the Fisher vector with 64 Gaussians.
A multi-layer perceptron (MLP) of three layers,
with a middle hidden layer of 100 nodes, was are trained.

Results are shown in Table \ref{tab:result}.
We compare the proposed TS feature with the original iDT feature and other recent methods.
Skip 2 version of TS feature doesn't perform well, showing that
we need to take care about parameter tuning for a better performance.
Exploring the effects of parameters (skipping, $M, K$, and $L$) is an important part of our future work.

By comparing with other recent methods, our TS feature outperforms the original iDT, and is better than
most of other methods, even without any appearance information of the scene.
We are now planning to validate how the proposed TS feature can be combined with other methods, including
deep learning approaches, for improving the performance.

\begin{table}[t]
\caption{Comparison of results on UFC50.}
\label{tab:result}
\centering

\begin{tabular}{l|c}
    & accuracy \\ \hline
Wang+2013 (DT) \cite{Wang2013} & 83.6 \\
Kataoka+2015 \cite{Kataoka2015} & 84.5 \\
Beaudry+2016 \cite{Beaudry2016} & 88.3 \\
TS skip2 (ours)  & \textbf{89.4} \\
Li+2016 \cite{Li2016} & 90.3 \\
Wang\&Schmid 2013 (iDT) \cite{iDT} & 91.7 \\
Peng+2016 \cite{Peng2016} & 92.3 \\
Yang+2017 \cite{Yang2017} & 92.4 \\
Lan+2015 \cite{Lan2015b} & 93.8 \\
Lan+2015 \cite{Lan2015} & 94.4 \\
Xu+2017 \cite{Xu2017} & 94.8 \\
TS (ours)  & \textbf{95.0} \\
Duta+2017 \cite{Duta2017a} & 97.8 \\
\end{tabular}
\end{table}

Recent work of action recognition uses more larger datasets,
such as UCF101 \cite{Soomro2012} and HMDB51 \cite{Kuehne2011}.
Tables \ref{tab:resultUCF101} and \ref{tab:resultHMDB51} show results.
For UCF101, the proposed TS feature is better than other methods before 2017,
but the recent methods presented in 2017 benefit clearly from the recent progress on deep learning.
For HMDB, however, our method outperforms all the deep learning-based methods by a clear margin, which is more than 5\%.
This is very surprising because our shallow method uses only the training sets provided, while
the recent method \cite{Carreira2017} uses more larger datasets
for training deep models with the help of feature transfer.

This results may indicate that CNN models used for recent activity recognition works
might not be as good as for image recognition. Features generated by CNN layers are completely
different from the TS features presented in this paper. A potential future work is to seek a deep model to
compute features from a batch of trajectory, not from pixel values or flows.

\begin{table}[t]
\caption{Comparison of results on UFC101.}
\label{tab:resultUCF101}
\centering

\begin{tabular}{l|c} & accuracy \\ \hline
Somroo+ 2012 \cite{Soomro2012} & 43.9 \\
Wang \& Schmid 2013 \cite{iDT} & 85.9\\
Wang+ 2015 \cite{Wang2015} & 88.0\\
Simonyan+ 2014 \cite{Simonyan2014} & 88.0 \\
TS (ours) & \textbf{88.6} \\
Kar+ 2017 \cite{Kar2017} & 93.2 \\
Duta+ 2017 \cite{Duta2017a} & 94.3 \\
Wang+ 2017 \cite{Wang2016} & 94.6 \\
Feichtenhofer+ 2017 \cite{Feichtenhofer2017} & 94.9 \\
Lan+ 2017 \cite{Lan2017} & 95.3 \\
Carreira \& Zisserman 2017 \cite{Carreira2017} & 97.9 \\

\end{tabular}
\end{table}

\begin{table}[t]
\caption{Comparison of results on HMDB51.}
\label{tab:resultHMDB51}
\centering

\begin{tabular}{l|c} & accuracy \\ \hline
Simonyan+ 2014  \cite{Simonyan2014} & 59.4 \\
Wang \& Schmid 2013 (iDT) \cite{iDT} & 61.7 \\
Kar+ 2017 \cite{Kar2017} & 66.9 \\
Wang+ 2017 \cite{Wang2017} & 68.9\\
Wang+ 2016 \cite{Wang2016} & 69.4\\
Feichtenhofer+ 2016 \cite{Feichtenhofer2016} & 70.3 \\
Feichtenhofer+ 2017 \cite{Feichtenhofer2017} & 72.2 \\
Duta+ 2017 \cite{Duta2017b} & 73.1 \\
Lan+ 2017 \cite{Lan2017} & 75.0 \\
Carreira \& Zisserman 2017 \cite{Carreira2017} & 80.2 \\
TS (ours) & \textbf{85.4} \\
\end{tabular}
\end{table}

\section*{Acknowledgments}

This work was supported in part by JSPS KAKENHI grant number JP16H06540.


\begin{thebibliography}{10}

\bibitem{Beaudry2016}
Cyrille Beaudry, Renaud P{\'{e}}teri, and Laurent Mascarilla.
\newblock {An efficient and sparse approach for large scale human action
  recognition in videos}.
\newblock {\em Machine Vision and Applications}, 27(4):529--543, may 2016.

\bibitem{Carreira2017}
Joao Carreira and Andrew Zisserman.
\newblock {Quo Vadis, Action Recognition? A New Model and the Kinetics
  Dataset}.
\newblock In {\em 2017 IEEE Conference on Computer Vision and Pattern
  Recognition (CVPR)}, pages 4724--4733. IEEE, jul 2017.

\bibitem{HOG}
Navneet Dalal and Bill Triggs.
\newblock Histograms of oriented gradients for human detection.
\newblock In {\em Proceedings of the 2005 IEEE Computer Society Conference on
  Computer Vision and Pattern Recognition (CVPR'05) - Volume 1 - Volume 01},
  CVPR '05, pages 886--893, Washington, DC, USA, 2005. IEEE Computer Society.

\bibitem{MBH}
Navneet Dalal, Bill Triggs, and Cordelia Schmid.
\newblock Human detection using oriented histograms of flow and appearance.
\newblock In {\em Proceedings of the 9th European Conference on Computer Vision
  - Volume Part II}, ECCV'06, pages 428--441, Berlin, Heidelberg, 2006.
  Springer-Verlag.

\bibitem{Duta2017a}
Ionut~C. Duta, Bogdan Ionescu, Kiyoharu Aizawa, and Nicu Sebe.
\newblock {Spatio-Temporal VLAD Encoding for Human Action Recognition in
  Videos}.
\newblock In {\em International Conference on Multimedia Modeling MMM 2017},
  pages 365--378. Springer, Cham, 2017.

\bibitem{Duta2017b}
Ionut~Cosmin Duta, Bogdan Ionescu, Kiyoharu Aizawa, and Nicu Sebe.
\newblock {Spatio-Temporal Vector of Locally Max Pooled Features for Action
  Recognition in Videos}.
\newblock In {\em 2017 IEEE Conference on Computer Vision and Pattern
  Recognition (CVPR)}, pages 3205--3214. IEEE, jul 2017.

\bibitem{Feichtenhofer2017}
Christoph Feichtenhofer, Axel Pinz, and Richard~P. Wildes.
\newblock {Spatiotemporal Multiplier Networks for Video Action Recognition}.
\newblock In {\em 2017 IEEE Conference on Computer Vision and Pattern
  Recognition (CVPR)}, pages 7445--7454. IEEE, jul 2017.

\bibitem{Feichtenhofer2016}
Christoph Feichtenhofer, Axel Pinz, and Andrew Zisserman.
\newblock {Convolutional Two-Stream Network Fusion for Video Action
  Recognition}.
\newblock In {\em 2016 IEEE Conference on Computer Vision and Pattern
  Recognition (CVPR)}, pages 1933--1941. IEEE, jun 2016.

\bibitem{Herath2017}
Samitha Herath, Mehrtash Harandi, and Fatih Porikli.
\newblock {Going Deeper into Action Recognition: A Survey}.
\newblock {\em Image and Vision Computing}, pages 1--18, feb 2017.

\bibitem{Kar2017}
Amlan Kar, Nishant Rai, Karan Sikka, and Gaurav Sharma.
\newblock {AdaScan: Adaptive Scan Pooling in Deep Convolutional Neural Networks
  for Human Action Recognition in Videos}.
\newblock In {\em 2017 IEEE Conference on Computer Vision and Pattern
  Recognition (CVPR)}, pages 5699--5708. IEEE, jul 2017.

\bibitem{Kataoka2015}
Hirokatsu Kataoka, Yoshimitsu Aoki, Kenji Iwata, and Yutaka Satoh.
\newblock {Evaluation of Vision-Based Human Activity Recognition in Dense
  Trajectory Framework}.
\newblock In {\em International Symposium on Visual Computing, Advances in
  Visual Computing}, pages 634--646. Springer, Cham, 2015.

\bibitem{Kuehne2011}
H.~Kuehne, H.~Jhuang, E.~Garrote, T.~Poggio, and T.~Serre.
\newblock {HMDB: A large video database for human motion recognition}.
\newblock In {\em Proceedings of the IEEE International Conference on Computer
  Vision}, pages 2556--2563. IEEE, nov 2011.

\bibitem{Lan2015b}
Zhenzhong Lan, Xuanchong Li, Ming Lin, and Alexander~G. Hauptmann.
\newblock {Long-short Term Motion Feature for Action Classification and
  Retrieval}.
\newblock Technical report, CoRR abs/1502.04132, 2015.

\bibitem{Lan2015}
Zhenzhong Lan, Ming Lin, Xuanchong Li, Alexander~G. Hauptmann, and Bhiksha Raj.
\newblock {Beyond Gaussian Pyramid: Multi-skip Feature Stacking for action
  recognition}.
\newblock {\em Proceedings of the IEEE Computer Society Conference on Computer
  Vision and Pattern Recognition}, 07-12-June-2015:204--212, 2015.

\bibitem{Lan2017}
Zhenzhong Lan, Yi~Zhu, Alexander~G. Hauptmann, and Shawn Newsam.
\newblock {Deep Local Video Feature for Action Recognition}.
\newblock {\em IEEE Computer Society Conference on Computer Vision and Pattern
  Recognition Workshops}, 2017-July:1219--1225, 2017.

\bibitem{Li2016}
Qingwu Li, Haisu Cheng, Yan Zhou, and Guanying Huo.
\newblock {Human Action Recognition Using Improved Salient Dense Trajectories}.
\newblock {\em Computational Intelligence and Neuroscience}, 2016:1--11, 2016.

\bibitem{Matikainen2009}
Pyry Matikainen, Martial Hebert, and Rahul Sukthankar.
\newblock {Trajectons: Action recognition through the motion analysis of
  tracked features}.
\newblock In {\em 2009 IEEE 12th International Conference on Computer Vision
  Workshops, ICCV Workshops 2009}, pages 514--521. IEEE, sep 2009.

\bibitem{Matsui2017}
Kenji Matsui, Toru Tamaki, Bisser Raytchev, and Kazufumi Kaneda.
\newblock {Trajectory-Set Feature for Action Recognition}.
\newblock {\em IEICE Transactions on Information and Systems},
  E100-D(8):1922--1924, 2017.

\bibitem{Peng2016}
Xiaojiang Peng, Limin Wang, Xingxing Wang, and Yu~Qiao.
\newblock {Bag of visual words and fusion methods for action recognition:
  Comprehensive study and good practice}.
\newblock {\em Computer Vision and Image Understanding}, 150:109--125, 2016.

\bibitem{UCF50}
Kishore~K. Reddy and Mubarak Shah.
\newblock Recognizing 50 human action categories of web videos.
\newblock {\em Mach. Vision Appl.}, 24(5):971--981, July 2013.

\bibitem{FisherVector}
Jorge S\'{a}nchez, Florent Perronnin, Thomas Mensink, and Jakob Verbeek.
\newblock Image classification with the fisher vector: Theory and practice.
\newblock {\em Int. J. Comput. Vision}, 105(3):222--245, December 2013.

\bibitem{Simonyan2014}
Karen Simonyan and Andrew Zisserman.
\newblock {Two-Stream Convolutional Networks for Action Recognition in Videos}.
\newblock In Z~Ghahramani, M~Welling, C~Cortes, N~D Lawrence, and K~Q
  Weinberger, editors, {\em Advances in Neural Information Processing Systems
  27}, pages 568--576. Curran Associates, Inc., 2014.

\bibitem{Soomro2012}
Khurram Soomro, Amir~Roshan Zamir, and Mubarak Shah.
\newblock {UCF101: A Dataset of 101 human actions classes from videos in the
  wild}.
\newblock Technical Report November, CRCV-TR-12-01, 2012.

\bibitem{Tran2015}
Du~Tran, Lubomir Bourdev, Rob Fergus, Lorenzo Torresani, and Manohar Paluri.
\newblock {Learning Spatiotemporal Features with 3D Convolutional Networks}.
\newblock In {\em 2015 IEEE International Conference on Computer Vision
  (ICCV)}, pages 4489--4497. IEEE, dec 2015.

\bibitem{DT}
Heng Wang, A.~Klaser, C.~Schmid, and Cheng-Lin Liu.
\newblock Action recognition by dense trajectories.
\newblock In {\em Proceedings of the 2011 IEEE Conference on Computer Vision
  and Pattern Recognition}, CVPR '11, pages 3169--3176, Washington, DC, USA,
  2011. IEEE Computer Society.

\bibitem{Wang2013}
Heng Wang, Alexander Kl{\"a}ser, Cordelia Schmid, and Cheng-Lin Liu.
\newblock Dense trajectories and motion boundary descriptors for action
  recognition.
\newblock {\em International Journal of Computer Vision}, 103(1):60--79, 2013.

\bibitem{iDT}
Heng Wang and Cordelia Schmid.
\newblock Action recognition with improved trajectories.
\newblock In {\em Proceedings of the 2013 IEEE International Conference on
  Computer Vision}, ICCV '13, pages 3551--3558, Washington, DC, USA, 2013. IEEE
  Computer Society.

\bibitem{Wang2015}
Limin Wang, Yu~Qiao, and Xiaoou Tang.
\newblock {Action recognition with trajectory-pooled deep-convolutional
  descriptors}.
\newblock In {\em 2015 IEEE Conference on Computer Vision and Pattern
  Recognition (CVPR)}, pages 4305--4314. IEEE, jun 2015.

\bibitem{Wang2016}
Limin Wang, Yuanjun Xiong, Zhe Wang, Yu~Qiao, Dahua Lin, Xiaoou Tang, and Luc
  van Gool.
\newblock {Temporal segment networks: Towards good practices for deep action
  recognition}.
\newblock In {\em European Conference on Computer Vision ECCV 2016}, volume
  9912 LNCS, pages 20--36. Springer, Cham, 2016.

\bibitem{Wang2017}
Yunbo Wang, Mingsheng Long, Jianmin Wang, and Philip~S. Yu.
\newblock {Spatiotemporal Pyramid Network for Video Action Recognition}.
\newblock In {\em 2017 IEEE Conference on Computer Vision and Pattern
  Recognition (CVPR)}, pages 2097--2106. IEEE, jul 2017.

\bibitem{Xu2017}
Zengmin Xu, Ruimin Hu, Jun Chen, Chen Chen, Huafeng Chen, Hongyang Li, and
  Qingquan Sun.
\newblock {Action recognition by saliency-based dense sampling}.
\newblock {\em Neurocomputing}, 236:82--92, 2017.

\bibitem{Yang2017}
Yang Yang, De-chuan Zhan, Ying Fan, Yuan Jiang, and Zhi-hua Zhou.
\newblock {Deep Learning for Fixed Model Reuse}.
\newblock In {\em Proceedings of the 31st AAAI Conference on Artificial
  Intelligence (AAAI17)}, San Francisco, 2017.

\end{thebibliography}
\end{document}